\documentclass[sigconf]{acmart}

\usepackage{multirow}
\usepackage{pifont}
\newcommand{\cmark}{\ding{51}}%
\usepackage{multirow}
\usepackage{hhline}
\usepackage{hyperref}
\usepackage{subfigure}
\usepackage{verbatim}

 \newcommand{\squishlist}{
	\begin{list}{$\bullet$}
		{ \setlength{\itemsep}{0pt}
			\setlength{\parsep}{3pt}
			\setlength{\topsep}{3pt}
			\setlength{\partopsep}{0pt}
			\setlength{\leftmargin}{1.5em}
			\setlength{\labelwidth}{1em}
			\setlength{\labelsep}{0.5em} } }
	
	\newcommand{\squishlisttwo}{
		\begin{list}{$\bullet$}
			{ \setlength{\itemsep}{0pt}
				\setlength{\parsep}{0pt}
				\setlength{\topsep}{0pt}
				\setlength{\partopsep}{0pt}
				\setlength{\leftmargin}{2em}
				\setlength{\labelwidth}{1.5em}
				\setlength{\labelsep}{0.5em} } }
		
		\newcommand{\squishend}{
	\end{list}  }

\AtBeginDocument{%
  \providecommand\BibTeX{{%
    \normalfont B\kern-0.5em{\scshape i\kern-0.25em b}\kern-0.8em\TeX}}}

\copyrightyear{2020}
\acmYear{2020}
\setcopyright{acmcopyright}
\acmConference[CIKM '20] {The 29th ACM International Conference on Information and Knowledge Management}{October 19--23, 2020}{Virtual Event, Ireland}
\acmBooktitle{The 29th ACM International Conference on Information and Knowledge Management (CIKM '20), October 19--23, 2020, Virtual Event, Ireland}
\acmPrice{15.00}
\acmDOI{10.1145/3340531.3411990}
\acmISBN{978-1-4503-6859-9/20/10}
\begin{document}
\fancyhead{}

\title{SWE2: SubWord Enriched and Significant Word Emphasized Framework for Hate Speech Detection}

\author{Guanyi Mou, Pengyi Ye, Kyumin Lee}
\affiliation{
	\institution{Worcester Polytechnic Institute}
}
\email{{gmou, pye3, kmlee}@wpi.edu}	






\begin{abstract}
Hate speech detection on online social networks has become one of the emerging hot topics in recent years. With the broad spread and fast propagation speed across online social networks, hate speech makes significant impacts on society by increasing prejudice and hurting people. Therefore, there are aroused attention and concern from both industry and academia. In this paper, we address the hate speech problem and propose a novel hate speech detection framework called \emph{SWE2}, which only relies on the content of messages and automatically identifies hate speech. In particular, our framework exploits both word-level semantic information and sub-word knowledge. It is intuitively persuasive and also practically performs well under a situation with/without character-level adversarial attack. Experimental results show that our proposed model achieves 0.975 accuracy and 0.953 macro F1, outperforming 7 state-of-the-art baselines under no adversarial attack. Our model robustly and significantly performed well under extreme adversarial attack (manipulation of 50\% messages), achieving 0.967 accuracy and 0.934 macro F1.
\end{abstract}



\maketitle




\section{Introduction}
Hate speech, ``abusive speech targeting specific group characteristics'' \cite{warner2012detecting}, has long been causing annoying disturbance to many people's lives, in terms of misleading the trends, shaping bias and discrimination, aggregating and aggravating conflicts among different religious, gender, and racial groups, etc. With the rapid growth of online social networks, hate speech is spreading faster and affecting a larger population than before in human history across the world\footnote{\url{https://www.cfr.org/backgrounder/hate-speech-social-media-global-comparisons}}. Therefore, quickly and accurately identifying hate speech becomes crucial for keeping a harmonic and healthy online social environment, mitigating the possible conflicts, and protecting the diversity of our society. Hate speech detection is also helpful for public sentiment analysis and is useful as one of the pre-processing steps in content recommendation and chatterbot development \cite{badjatiya2017deep}.

Over the years, researchers have proposed various methods for detecting hate speech \cite{nobata2016abusive,djuric2015hate,badjatiya2019stereotypical,liu2019fuzzy,chowdhury2019arhnet,zhang2018detecting}, many of which have focused on feature engineering. New hand-crafted features were raised and checked from different perspectives to improve their overall performance. The content of hate speech was inevitably leveraged to generate features. The produced features vary from counting based features, sentiment, and semantic features to pre-trained word-level embedding and sentence-level embedding related features. However, the procedures of generating these features highly depend on two crucial presumptions: the sentences can be successfully tokenized into completely genuine atomized words, and these words can be recognized/categorized into bins. Thus, these methods/features are intuitively vulnerable in resisting character-level manipulation, which converts semantically significant known words to meaningless unknown words \cite{li2019textbugger}.

While in practice, it has been reported that intentionally or deliberately misspelled words as a kind of adversarial attacks are commonly adopted as a tool in manipulators' arsenal to evade detection. These manipulated words may vary in many ways but may not occur in normal natural language processing (NLP) related dictionaries \cite{warner2012detecting,nobata2016abusive}. In fact, it is not practical to generate a vocabulary that includes everything, as the atmosphere of online social networks is dynamic: new words/variations are emerging almost every day. Even legitimate users can sometimes accidentally make typos \cite{sproat2001normalization}. 
The prior hate speech detection methods may not handle these cases properly.

To address the mentioned problem, we propose \textbf{SWE2}, the \emph{\textbf{S}ub\textbf{W}ord \textbf{E}nriched and \textbf{S}ignificant \textbf{W}ord \textbf{E}mphasized} framework, which not only embraces the word-level semantics but also leverages sub-word information, to recognize typos/misspellings, resist character-level adversarial attacks and improve robustness and accuracy of hate speech detection. Our framework incorporates two types of subword embeddings: the phonetic-level embedding and the character-level embedding. In addition, we carefully designed an LSTM+attention based word-level feature extraction method, which extracts general content semantic information across the speech. For subword representations, we trained our domain-specific embeddings. While for word representations, we tested two pre-trained variations: the state-of-the-art generalized FastText embedding \cite{joulin2016fasttext}, and the latest and most advanced generalized BERT embedding \cite{devlin2019bert} to see which word representation complements our subword representations for hate speech detection.
With the combination of word-level and subword-level representations, our framework can achieve high performance and robustness with/without the character-level adversarial attack (i.e., intentionally manipulating some characters of a message to evade hate speech detection).

To sum up, the contributions of this paper are listed as follows:
\squishlist
    \item We proposed to incorporate two new vital proportions of sub-word information in hate speech detection: character-level information and phonetic-level information. Both of them are non-trivial and work as complementary to each other. To the best of our knowledge, we are the first to incorporate pronunciation information in hate speech detection domain and show it makes a non-trivial contribution.

    \item We designed a novel hate speech detection framework, which utilized CNNs for subword information extraction, LSTMs for word-level information extraction, and attention mechanisms with statistical MAXes and MEANs for better feature generation. In the word-level information extraction, we compared FastText and BERT to see which one contributes more in the framework.

    \item We investigated the performance of our model without and with a black-box adversarial attack. We showed our model outperformed 7 state-of-the-art baselines, achieving minor reduction even under very extreme attack (i.e., manipulation of 100\% messages in the test set), indicating robustness of our model.

    \item Our model only relies on the content of a hate speech message, not requiring other side information such as a user profile or message propagation status in the network. Therefore, it is more efficient than some of the prior methods in terms of feature utilization and prediction time.
\squishend

\begin{figure*}
    \centering
    \includegraphics[width=0.6\linewidth,height=1.2in]{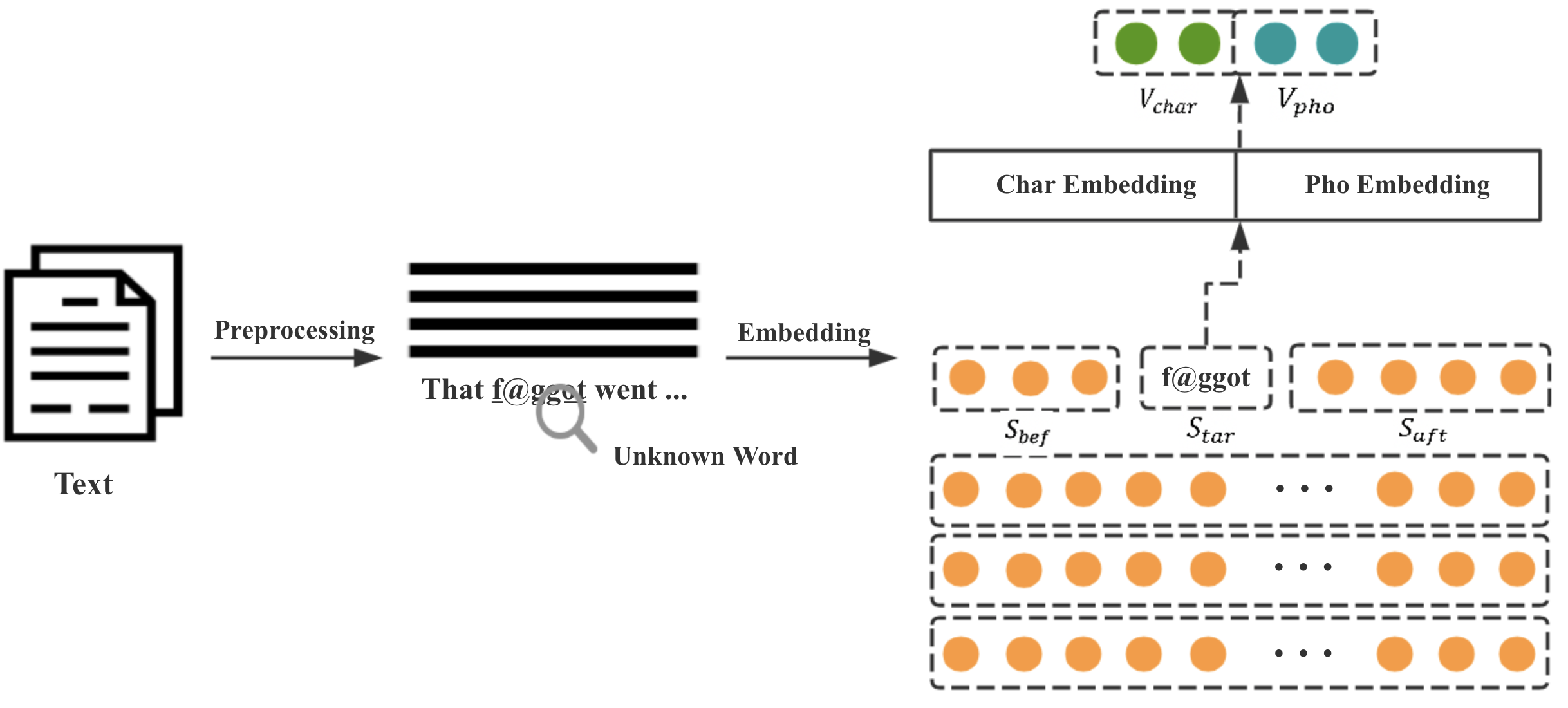}
    \vspace{-10pt}
    \caption{Process of extracting word-level and subword-Level embeddings in our framework.}
    \label{fig:OverallFramework}
    \vspace{-16pt}
\end{figure*}

\vspace{-8pt}
\section{Related Works}
\label{sec:RelatedWorks}
\subsection{Hate Speech Detection}
In the early literature \cite{davidson2017automated}, hate speech was defined as ``the language used to express hatred against a specific target group or to be demeaning, insulting, or insult the team members''. Mai ElSherief \cite{elsherief2018hate} specified hate speech into two classes: (1) directed hate -- hate language towards a specific individual or entity; and (2) generalized hate -- hate language towards a general group of individuals who share a common protected characteristic.

Despite the existing work dedicated to detecting hate speech \cite{schmidt2017survey}, hate speech detection is still challenging in the NLP domain. The main reason is due to linguistic diversity and words' semantic ambiguity.
Even though there have been several public corpora and automatic linguistic classifiers focusing on hate speech detection, the limitation and weakness of the existing classifiers based on text features are apparent. Intentional typos/character-level manipulation made by hate speech posters can easily avoid the prior hate speech detection methods based on texture features.

Ousidhoum et al. \cite{ousidhoum2019multilingual}, Davidson et al. \cite{davidson2017automated}, Waseem and Hovy \cite{waseem2016hateful}, and Elsherief et al. \cite{elsherief2018hate} released their annotated hate speech datasets in public. We made use of the latter three datasets in our research.  Mathew et al. \cite{mathew2019thou} and Chung et al. \cite{chung2019conan} provided counter speech datasets for better analysis of hate speech.

There are also papers focusing on in-depth analysis of hate speech: Warner and Hirschberg \cite{warner2012detecting} did an overall analysis of hate speech detection. Waseem \cite{waseem2016you} wrote about the annotator's influence on hate speech detection, showing expert annotations contributed to a better detection rate. 
Arango et al. \cite{arango2019hate} analyzed a model validation problem of hate speech detection.

For classification tasks, Nobata et al. \cite{nobata2016abusive} tried various types of features and reported informative results. Recent papers leveraged CNNs, LSTMs and attention mechanisms for better detection results \cite{badjatiya2017deep,gamback2017using,liu2019fuzzy,chowdhury2019arhnet,badjatiya2019stereotypical,zhang2018detecting}. Djuric et al. \cite{djuric2015hate} experimented on using paragraph-level embeddings for hate speech detection.

Our approach also incorporates LSTM, CNNs, and an attention mechanism. However, our approach differs from the prior works in the following ways:
\squishlist
    \item First, we use word-level, character-level, and phonetic-level embeddings for hate speech detection, whereas the prior works only focused on one-level representation.
    \item Second, we apply different techniques for different parts of our framework to best utilize each method's advantage (i.e., CNNs for subword information extraction, and LSTMs for word-level information extraction), whereas the prior works used LSTMs and CNNs with simple concatenation or sequential addition.
    \item While other state-of-the-art methods relied on spelling correcting tools such as \cite{gong2019context} or spellchecker\footnote{\url{https://norvig.com/spell-correct.html}} for word recovery (i.e., to fix typos), our framework is focusing on direct prediction without the word recovery (and without using spelling correcting tools) so that it can avoid possible new bias/error caused by the spelling correcting methods.
\squishend

We discuss why and how our ideas of utilizing LSTMs are better than the prior approaches in Section \ref{sec:CPLSTM}.

\vspace{-10pt}
\subsection{Adversarial Attack}
\label{subsec:Adversarial}
In the black-box attack, attackers do not know details of a detection framework, thus only generates the best possible guess to avoid exposure. In this paper, we focus on a black-box attack, especially, called a character-level adversarial attack. There is much clear evidence in character-level manipulations in online social networks. For instance, when a word\footnote{\textbf{\label{footn:Disclaimer}It is crucial to note that this paper contains hate speech examples (in Section \ref{subsec:Adversarial}, \ref{subsec:Manipulate}, \ref{subsec:CaseStudy}, and Table~\ref{tab:BugExample}), which may be offensive to some readers. They do not represent views of the authors. We tried to make a balance between showing less number of hate speech examples and sharing transparent information.}} `nigger' has a misspelling and is changed to `n1gger' or `nigga' which cannot be recognized by most word embedding models.
There are also systematic methods for generating character-level manipulations such as \cite{li2019textbugger,gong2019context}. They reported general methods for attacking different existing frameworks and claimed success in almost all experiments.

Serra et al. \cite{serra2017class} analyzed how out-of-vocabulary words can affect classification errors in detecting hate speech. Later on, Grondahl et al. \cite{grondahl2018all} reported that character models are much more resistant to simple text-transformation attack against hate speech classification. Inspired by that, our framework incorporates the subword information to detect hate speech containing character-level manipulation and achieves better prediction accuracy than the prior methods.

\vspace{-5pt}
\section{Background: Embedding Methods}
\label{sec:Background}
\noindent\textbf{Sentence-Level Embedding.}
Many existing works search for a higher level of text encodings, such as sentence-level or paragraph-level. Researchers proposed innovative ways to produce them \cite{cer2018universal,howard2018universal,subramanian2018learning,conneau2017supervised,logeswaran2018an,kiros2015skip,ruckle2018concatenated}.

\smallskip
\noindent\textbf{Word-Level Semantic Embedding.}
There are many state-of-the-art semantic embedding methods such as context-independent Word2Vec \cite{mikolov2013efficient}, and GloVe combining counting based vectors \cite{pennington2014glove}, character-based embeddings such as FastText \cite{joulin2016fasttext}, shallow bidirectional LSTM generated context-related ELMo \cite{peters2018deep}, and the most recent transformer generated BERT \cite{devlin2019bert}. In our research, we utilized each of FastText and BERT to see which one performs well in our framework for hate speech detection.

\smallskip
\noindent\textbf{Character-Level Embedding.}
The effectiveness of character-level information in the NLP domain was described in \cite{ling2015character}. Similarly, a variety of works exploring the subword-based representations in this domain emerged consistently from several perspectives, such as  part-of-speech tagging \cite{santos2014learning}, parsing \cite{ballesteros2015improved}, and normalization\cite{chrupala2014normalizing}. \cite{chen2015joint} explored such linguistic patterns in Chinese and proposed a character-enhanced word embedding model.

It is reported that many traditional word representations ignore the morphology of words, by assigning a distinct vector to each word. Bojanowski et al. \cite{bojanowski2017enriching} note: ``This is a limitation, especially for languages with large vocabularies and many rare words.''. Inspired by their discovery and effort in emphasizing such information, we explore the insight of words by incorporating the character-level embedding in our research.

\smallskip
\noindent\textbf{Phonetic-Level Embedding.}
Nowadays, phonetic recognition gains its popularity with the evolutions of hardware and software. CMU provided pronouncing dictionary\footnote{\url{http://www.speech.cs.cmu.edu/cgi-bin/cmudict\#phones}} to translate characters of each word into phonetic symbols. The phoneme distribution and syllable structure of words in this dictionary have been explored by \cite{yang2016phoneme} and compared with the results obtained from the Buckeye Corpus \cite{yang2012reduction}. Peng et al. \cite{peng2019phonetic} experimented with the strength of the pronunciation features of the text over sentiment analysis in Chinese by enriching the text representations. Based on the inspiration, we propose to incorporate the phonetic embedding method into our framework for hate speech detection, and further propose and develop a way to extract symbolic/phonetic representation of unknown words to learn their phonetic embeddings. The detailed approach of extracting phonetic embedding of unknown words is described in Section~\ref{sec:embedding}.

\begin{figure*}
    \centering
    \includegraphics[width=0.6\linewidth,height=1.6in]{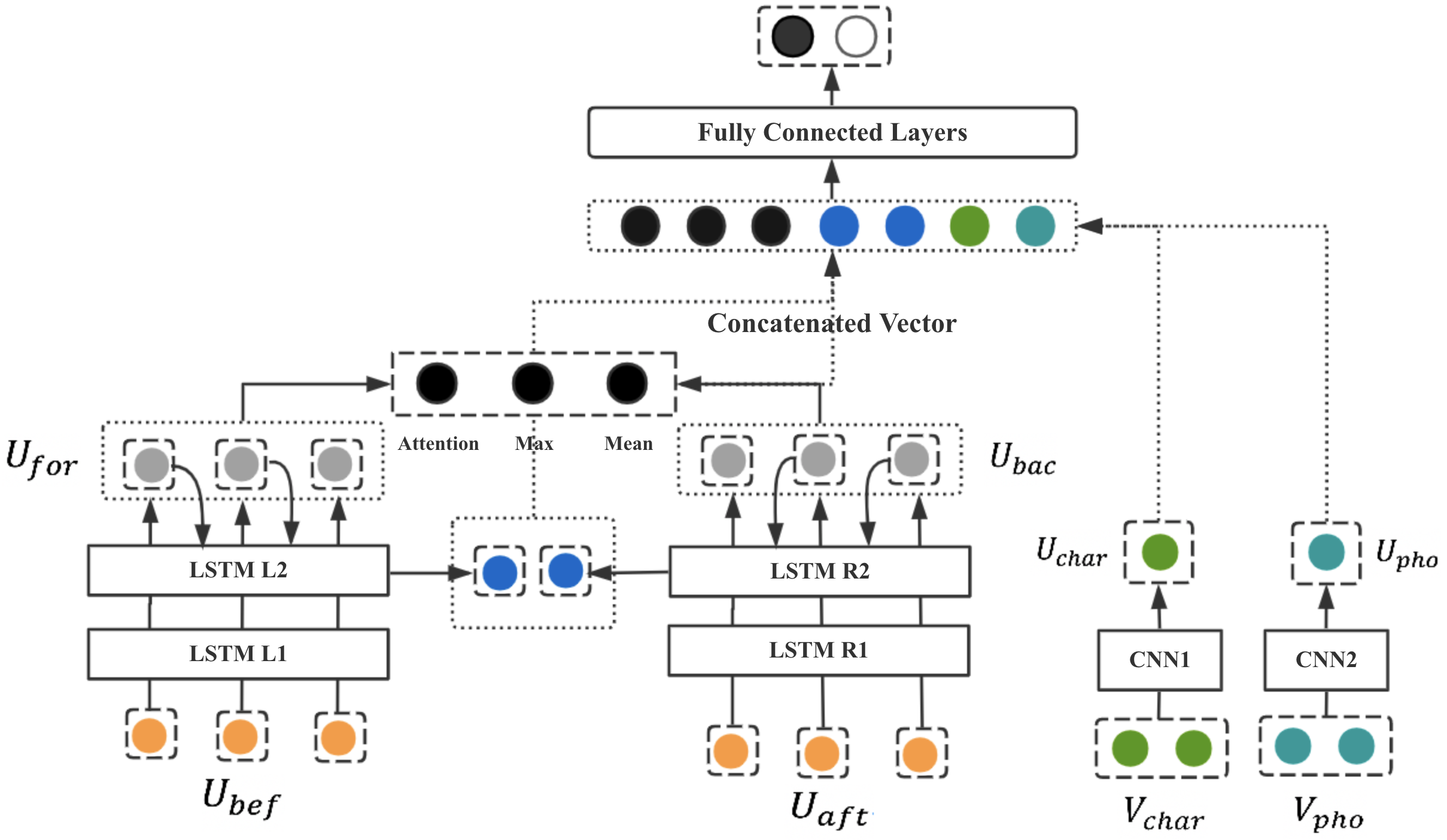}
    \vspace{-10pt}
    \caption{The overall framework.}
    \label{fig:SpecifiedFramework}
    \vspace{-10pt}
\end{figure*}

\vspace{-5pt}
\section{Our Framework}
\label{sec:CPLSTM}

We introduce our novel hate speech detection framework, \textbf{SWE2} (pronounced as `sweet'), which only relies on the text content of a message, identifies most essential words, and extracts their surrounding information to predict whether it is hate speech or not. Figure~\ref{fig:OverallFramework} shows the process of extracting word-level and subword-level embeddings in our framework, and Figure~\ref{fig:SpecifiedFramework} shows the overall framework which uses these embeddings as a part of input, and outputs the final prediction.

\subsection{Task and Procedure}
\label{subsec:Task&Procedures}

\smallskip
\noindent\textbf{Task.} Given a text message/speech, the framework ought to make a prediction of whether such delivered speech is hate speech or not.

\smallskip
\noindent\textbf{Cleaning.} We clean the messages as follows: we intentionally remove all sentence punctuation and make the message to lower case since what we care about in hate speech is the content itself.

\smallskip
\noindent\textbf{Tokenization and redundant information handling.} Then, we tokenize the message, remove special characters for post tags and only keep their content, replace mentions of usernames with ``USER'' and links with ``URL'', and feed the result to our network. Assuming the message is a tweet obtained from Twitter. However, our framework can handle any other text messages from other web sites with minimum customization.

\smallskip
\noindent\textbf{Most significant word recognition.} Given the tokenized word (string) sequence, we aim to identify the most significant word (called \emph{target} word). We first use VADER \cite{hutto2014vader} for searching the most sentimentally strong word. If all words are sentimentally contributing similarly, we will compare each word with possible hate speech words in the given dictionary \cite{elsherief2018hate} to see whether hate speech words are having at most two-character difference compared with the word in the tokenized word sequence. We can quickly achieve such a goal by implementing a simple longest common subsequence method. If we find no similar word from the sequence, then the framework will randomly assign a word as the \emph{target} word. Section~\ref{sec:targetword} describes why we care about identifying a \emph{target} word of each message and why our framework pays extra attention to the \emph{target} word.

\smallskip
\noindent\textbf{Splitting and Embedding.}
Given the \emph{target} word in a sentence/ message, we split the original sentence $S_{Ori}$ into three parts, namely part before \emph{target} word $S_{Bef}$, \emph{target} word $S_{Tar}$, and the part after \emph{target} word $S_{Aft}$.
\vspace{-5pt}
\begin{equation}
    S_{Ori} = [S_{Bef}, S_{Tar}, S_{Aft}]
\end{equation}

We use the character-level and phonetic-level embeddings trained by ourselves to represent the \emph{target} word $S_{Tar}$, deriving representations of $V_{Char}$ and $V_{Pho}$ as shown in Figure~\ref{fig:OverallFramework}.
\vspace{-5pt}
\begin{equation}
    V_{Char} = EmbC(S_{Tar})
\end{equation}
\begin{equation}
    V_{Pho} = EmbP(S_{Tar})
\end{equation}

We feed these two matrices to separate CNNs, trying to fetch important information of the \emph{target} word as shown in Figure~\ref{fig:SpecifiedFramework}.
\vspace{-5pt}
\begin{equation}
    U_{Char} = CNN1(V_{Char})
\end{equation}
\begin{equation}
    U_{Pho} = CNN2(V_{Pho})
\end{equation}

We use word-level embedding methods to represent each word in $S_{Bef}$ and $S_{Aft}$ separately, stacking word vectors together and receive representations of two matrices $U_{Bef} \in R^2$ and $U_{Aft} \in R^2$.
\vspace{-5pt}
\begin{equation}
    U_{Bef} = EmbW(S_{Bef})
\end{equation}
\begin{equation}
    U_{Aft} = EmbW(S_{Aft})
\end{equation}

Now two matrices representing the part of the message before and after the \emph{target} word are fed to two separate LSTM models, to extract useful information as shown in Figure~\ref{fig:SpecifiedFramework}. Inspired by the ELMo design \cite{peters2018deep}, which had two layers of LSTM to learn syntax and grammar of the sentence in the first layer and to address semantic meanings of words and disambiguation in the second layer, each of our two LSTM models has two layers. We call the first LSTM model as forward LSTM and the second one as backward LSTM because $U_{Bef}$ is fed to forward LSTM, and $U_{Aft}$ is reversed (i.e. reversing the order of word vectors of $U_{Aft}$), and then is fed to backward LSTM. In this way, the last outputs of the two LSTMs can be treated as the prediction of the \emph{target} word, while the other outputs of the two LSTMs can be seen as global side information. All of the outputs from the second layers of the two LSTMs are collected and formed as two matrices $U_{For} \in R^2$ and $U_{Bac} \in R^2$.
\vspace{-5pt}
\begin{equation}
    U_{For} = LSTM_{Forward}(U_{Bef})
\end{equation}
\begin{equation}
    U_{Bac} = LSTM_{Backward}(Reverse(U_{Aft}))
\end{equation}

Now we split each output matrix into two parts: the last outputs of two LSTMs can easily be explained as the predicted vector representation of the \emph{target} word (i.e., $U_{ForLast}$ and $U_{BacLast}$), so we separate them from the other outputs (i.e., $U_{ForRest}$ and $U_{BacRest}$).
\vspace{-5pt}
\begin{equation}
    U_{For} = U_{ForLast} \oplus U_{ForRest}
\end{equation}
\begin{equation}
    U_{Bac} = U_{BacLast} \oplus U_{BacRest}
\end{equation}
where $\oplus$ denotes concatenation.

Eventually we form the global information $U_{Glo} \in R^2$, and also the focused local representation $U_{Loc} \in R^1$. Notice that global information is everything except the \emph{target} word, while the focused local representation is only about the \emph{target} word.
\vspace{-5pt}
\begin{equation}
    U_{Glo} = U_{ForRest} \oplus U_{BacRest}
\end{equation}
\begin{equation}
    U_{Loc} = U_{ForLast} \oplus U_{BacLast} \oplus U_{Char} \oplus U_{Pho}
\end{equation}

Now the global information is too large and may contain much redundancy, so we only use self-attention, max and mean information extracted from it and concatenate them, namely $U_{Glo2} \in R^1$
\vspace{-5pt}
\begin{equation}
    U_{Glo2} = Attn(U_{Glo}) \oplus Max(U_{Glo}) \oplus Mean(U_{Glo})
\end{equation}

Finally, we combine the global information and local information altogether, feed them to multiple fully connected layers, and then make the final prediction.
\vspace{-5pt}
\begin{equation}
    Pred(S_{Ori}) = argmax(MultiFC(U_{Glo2} \oplus U_{Loc}))
\end{equation}

\vspace{-5pt}
\subsection{Character-Level Manipulation for the Adversarial Attack}
\label{subsec:Manipulate}
In this subsection, we describe how to simulate the character-level adversarial attack. Due to the diversity of text combinations and varieties of typos, it is labor-consuming to manually collect large-scaled hate speech data, which contains deliberate typos as well as recovering their original perfect spelling. To the best of our knowledge, there is no existing publicly available large scale dataset, which includes both real-world hate speech with deliberate typos and the recovered ones.

So we turned to apply scalable simulation methods to generate spelling errors. We used \textit{TEXTBUGGER} framework \cite{li2019textbugger}, which can create utility-preserving adversarial texts against the state-of-the-art text classification systems effectively and efficiently, focusing on the \emph{target} word manipulation and mimicking evasion behavior of hate speech posters as we described them in Section~\ref{sec:targetword}.

According to \cite{li2019textbugger}, we could consider five types of bug generations (i.e., manipulation) inside one word (i.e., the \emph{target} word):
\squishlist
    \item Insertion: insert an extra space in the word
    \item Deletion: delete a character in the word
    \item Swapping: switch the position of two neighbor characters
    \item Sub-c: substitute a character with a similar character
    \item Sub-w: replace the word with its closest meaning neighbor
\squishend

From the description of hate speech on \textit{Wikipedia}\footnote{\url{https://en.wikipedia.org/wiki/Hate_speech}}, the domain of the hate speech is deeply nested in the offensive and hostile speech. However, the boundary of hate speech is so strict, so even the nearest word in semantics or word embedding representations can unlikely be hate speech. For instance, the most similar word\footnote{Here we used Word2Vec vector and measured with cosine similarity} of ``limey'', which is an insulting word for a British person, is ``yeasty'', which might be offensive in some scenarios but far from being hate speech. Therefore, in our experiments, we choose not to do any word-level manipulation to avoid any labeling bias introduced to the manipulated data. Based on that, \emph{sub-w} is not selected.

\begin{table}[t]
    \centering
    \small
    \caption{Examples of character-level manipulation.}
    \vspace{-10pt}
    \scalebox{0.85}{
    \begin{tabular}{c|c c|c c}\hline
        & \multicolumn{2}{c|}{Char}&   \multicolumn{2}{c}{Phonetic}\\ 
        \textbf{Method}&   Original&   Manipulated&    Original&   Manipulated\\ \hline
        \textbf{Swap}&   fucking&    fukcing&    limey&  liemy\\
        \textbf{Delete}& wigger& wiger&  coonass&    coonas\\
        \textbf{Sub-C}&  trash&  tr@sh&  nigger& neegeer\\ \hline
    \end{tabular}}
    \label{tab:BugExample}
    \vspace{-10pt}
\end{table}

Insertion is also not chosen for two reasons. First, inserting a space can be interpreted as a method of word-level manipulation, where one word is split into two words if we add whitespace inside of the word. Secondly, splitting one word into two words may severely impact the readability of the whole speech. Taking these scenarios into consideration, we use character-level \emph{sub-c}, \emph{deletion}, and \emph{swapping}
to keep the original meaning of a message for human readers. We show examples of the character-level manipulations in Table~\ref{tab:BugExample}. Sometimes hate speech posters do character-level manipulation with/without considering phonetic similarity. The examples consider both cases.

To generate the character-level manipulation, our attack system automatically searches for hate speech candidate words in the aforementioned hate speech words dictionary. Then, it selects one of them for the manipulation. If there was no hate speech candidate words, the manipulation happens in a randomly selected word.

To decide which attacking method among \emph{sub-c}, \emph{deletion}, and \emph{swapping} is the most effective for a given message, we used Universal Sentence Encoder \cite{cer2018universal}, a sentence-level embedding framework, to encode the whole message into a fixed-length vector. Then we used cosine similarity to compare the distance of the original message (before the manipulation) and the manipulated message (after the manipulation). We chose the attacking method that produces the longest distance among them.

\vspace{-5pt}
\section{Experiments}
\label{sec:Experiment}
\subsection{Data Collection}
To conduct experiments, we used some portion of three existing datasets (i.e., Waseem \cite{waseem2016hateful}, Davidson \cite{davidson2017automated} and HateLingo \cite{elsherief2018hate} datasets) and collected legitimate tweets from Twitter as follows:

    \smallskip\noindent\textbf{Waseem dataset \cite{waseem2016hateful}} includes 17,325 tweets which were manually labeled into sexism, racism, offensive, and neither. The labels of the messages were automatically identified, and the reliability and consistency of labels were manually investigated and verified. As we are aware of the fact that offensive speeches do not necessarily lead to hate speech, we filtered out the offensive messages; however, we kept the sexism and racism as hate speech. Eventually, we fetched out 2,778 hate speech and 7,133 legitimate speech from the dataset.

    \smallskip\noindent\textbf{Davidson dataset \cite{davidson2017automated}} includes 24,783 tweets, consisting of offensive speech, hate speech, and neither.
        Similar to what we did in the previous dataset, we removed offensive speech, eventually using 1,294 hate speech and 3,925 legitimate messages.

    \smallskip\noindent\textbf{HateLingo dataset \cite{elsherief2018hate}} contains only hate speech messages crawled via Twitter Streaming API with specific keywords and hashtags defined by Hatebase\footnote{\url{https://www.hatebase.org/}}. To recognize the anti-hate tweets which may also contain hate speech terms, the authors cleaned the dataset by using Perspective API\footnote{\url{https://github.com/conversationai/perspectiveapi/blob/master/api_reference.md}}, and conducted manual checking during the experiment. As they only provided TweetID, we had to fetch the actual messages through Twitter API. We were able to collect 12,631 hate speech messages.

    \smallskip\noindent\textbf{Legitimate Messages}: To balance the proportion of hate speech and legitimate messages, we randomly collected 1\% real-time tweets by Twitter API, and then selected 800,000 tweets in English. To guarantee these tweets are legitimate messages, we followed thorough labeling process: filter out the messages which contain the aforementioned hate speech keywords or have negative sentiment scores by following the rule proposed in \cite{elsherief2018hate}. Finally, 72,457 messages were chosen, and we manually sampled 2,000 messages to see whether there is any hate speech message or not. All of them were legitimate. Therefore, we labeled them as legitimate messages. Dataset can be found at \url{https://web.cs.wpi.edu/~kmlee/data.html}.

Our dataset was carefully selected from various datasets to avoid possible bias in any single data collection and labeling method. We incorporated hate speech messages which contain \emph{target} words and those which do not. All of these efforts were made to avoid the possibilities that a model only learned to identify particular words or particular hashtags. As the world is evolving, word meanings can sometimes be ambiguous. A message containing certain words may not necessarily be absolute hate speech \cite{badjatiya2019stereotypical}. Overall, our dataset consists of 16,703 hate speech messages and 83,515 legitimate messages as presented in Table~\ref{tab:dataset}. Note that all of our datasets are related to Twitter. We chose these datasets as they are easy to trace and verify. However, our framework is not designed for only Twitter but designed for any online social system because it only requires a text message without any other additional information (e.g., user profile, social network, temporal/activity information) for the hate speech detection.

\begin{table}[t]
    \centering
    \small
    \caption{Dataset.}
    \vspace{-10pt}
    \scalebox{0.81}{
    \begin{tabular}{l r r}\hline
        \textbf{source}&    \textbf{|hate speech|} &  \textbf{|legitimate|}\\ \hline
         Waseem \cite{waseem2016hateful}&    2,778 & 7,133\\
         Davidson \cite{davidson2017automated}&  1,294 & 3,925\\
         HateLingo \cite{elsherief2018hate}& 12,631 & 0\\
         Legitimate& 0 & 72,457\\\hline
         Total & 16,703 & 83,515
    \end{tabular}}
    \label{tab:dataset}
    \vspace{-10pt}
\end{table}

\vspace{-5pt}
\subsection{Preprocessing}
\label{subsec:preprocessing}
We preprocessed the collected dataset in the following ways to provide cleaner data to our framework as well as to guarantee the capability of generalizing our model in any online social media:

\squishlist
    \item As mentioned in Section \ref{subsec:Task&Procedures}, we removed all punctuations and irrelevant characters. We converted all letters to lower case.
    \item Privacy information such as user mention (i.e., @username) was substituted with ``USER''. Domain-specific labels such as hashtags had their starting character removed (e.g., for Twitter, hashtags start with `\#') and content kept. Specific website links are kept anonymous as ``URL''. In this way, we guarantee the generalized ability of models.
\squishend

\begin{table*}[t]
    \centering
    \small
    \caption{Baseline Information.}
    \vspace{-10pt}
    \scalebox{0.85}{
    \begin{tabular}{l c | c c | l} \hline
        \multirow{2}{*}{Model}&  Domain&   Deep&    Machine&  \multirow{2}{*}{Description}\\
        &   Specific&  Learning&  Learning&  \\ \hline
        Davidson'17&    \cmark& &   \cmark& Linear SVC trained on a combination of useful handcrafted features. \\
        Text-CNN'14&    &   \cmark& &   CNN-based model with dynamic window size for text classification.\\
        Badjatiya'17&   \cmark& \cmark& & LSTM-based network designed for hate speech detection.\\
        Waseem'16&  \cmark& &   \cmark& Logistic regression model trained on n-gram counting-based features.\\
        Zhang'18&   \cmark& \cmark& &   CNN followed by GRU, then fully connected layers for classification.\\
        Fermi'19&   \cmark& \cmark& \cmark&   USE for sentence embedding, then SVM for classification.\\
        DirectBert'19&  &   \cmark& &   Pooling BERT embeddings into sentence embeddings, concatenate with MLP.\\

        \hline
    \end{tabular}}
    \label{tab:BaselineFeatures}
    \vspace{-10pt}
\end{table*}

\vspace{-5pt}
\subsection{Embeddings}
As we mentioned in the previous section, three types of embeddings were generated in our framework: phonetic-level embedding, character-level embedding, and word-level embedding.

\vspace{-5pt}
\subsubsection{\textbf{Phonetic-level embedding}}
\label{sec:embedding}
We trained our domain-specific phonetic-level embedding in the following way:
\begin{enumerate}
    \item \emph{\textbf{Data Collection.}} We randomly collected 80,000 tweets from Twitter API for training our embeddings and further guaranteed that these tweets do not have any overlap with the experiment dataset. These tweets then went through the same data preprocessing procedure described in Section \ref{subsec:preprocessing}. Among these tweets, there were 40,000+ unique words.
    \item \emph{\textbf{Known Word Pronunciation Conversion.}} We used the CMU pronouncing dictionary to translate characters in each word into phonetic symbols. Such a dictionary can only handle known words in a given fixed size vocabulary.
    \item \emph{\textbf{Unknown Word Pronunciation Prediction.}} To overcome the limitation of the CMU pronunciation dictionary and extract symbolic (phonetic) representation of unknown words, the known words' sequences of symbols were fed to an attentive LSTM model\footnote{\url{https://github.com/repp/big-phoney}} for training. Eventually, the model can predict any given unknown word's symbolic representation.
    \item \emph{\textbf{Embedding.}} Given any word's symbolic representation, we trained the embedding with the same design of Word2Vec CBOW method \cite{godin2019}. Thus eventually each word is embedded into a 2D matrix, with its height as the number of phonetic symbols and the width as the vector dimension.
\end{enumerate}

\vspace{-5pt}
\subsubsection{\textbf{Character-level embedding}}
We also trained a domain-specific character-level embedding in the following way:
\begin{enumerate}
    \item \emph{\textbf{Data Collection.}} We used the same dataset collected and used for the phonetic-level embedding.
    \item \emph{\textbf{Embedding.}} We trained our own character-level embedding model to directly predict embedding using the same design of Word2Vec CBOW method, thus getting a 2D-Matrix similar to phonetic-level embedding.
\end{enumerate}

\vspace{-5pt}
\subsubsection{\textbf{Word-level embedding}}
To understand which word-level embedding under our framework performs the best, we applied two popular word embedding models: FastText and BERT.
FastText is a character-based embedding model trained by large corpus of data, which is capturing more subword information. BERT \cite{devlin2019bert} is a context related model that achieved outstanding performance in many NLP tasks. BERT also used a word-piece tokenizer, which enabled it to capture subword information effectively and is thus intuitively less vulnerable to character-level adversarial attacks.

\vspace{-5pt}
\subsection{Baselines}
To compare the performance of our model against baselines, we chose the following \textbf{7} state-of-the-art baselines as shown in Table~\ref{tab:BaselineFeatures}:

    \smallskip\noindent\textbf{Davidson'17} \cite{davidson2017automated}: This model used the linear support vector classifier trained by TPO features (i.e., TF-IDF and POS), and other text features such as sentiment scores, readability scores and count indicators for \# of mentions and \# of hashtags, etc.

    \smallskip\noindent\textbf{Text-CNN'14} \cite{kim2014convolutional}: Inspired by Kim's work, we implemented Text-CNN for hate speech detection trained by using GloVe as default word embedding. This is a general purpose classification framework widely applied in many text classification tasks.

    \smallskip\noindent\textbf{Badjatiya'17} \cite{badjatiya2017deep}: It is a domain specific LSTM-based network for detecting hate speeches.

    \smallskip\noindent\textbf{Waseem'16} \cite{waseem2016hateful}: This baseline is a logistic regression model trained on the bag of words features.

    \smallskip\noindent\textbf{Zhang'18} \cite{zhang2018detecting}: The authors of this work proposed C-GRU model, which combines CNN and GRU to not only fit in small-scale data but also accelerate the whole progress. Such framework is domain specific to hate speech detection.

    \smallskip\noindent\textbf{Fermi'19} \cite{indurthi2019fermi}: \emph{Fermi} was proposed for Task 5 of SemEval-2019: HatEval: Multilingual Detection of Hate Speech Against Immigrants and Women on Twitter. The authors participated in the subtask A for English and ranked the first in the evaluation on the test set. Their model used pretrained Universal Encoder sentence embeddings for transforming the input, and SVM 
    for classification.

    \smallskip\noindent\textbf{Directly using BERT for sentence encoding (DirectBERT'19)}: As an additional baseline, we applied BERT in directly generating sentence encodings with two linear projection layers and dropout. We used REDUCE\_MEAN, which takes the average of the hidden state of the encoding layer on the time axis for pooling strategy. It maps word-piece embeddings to the whole sentence embedding.

\vspace{-10pt}
\subsection{Experiment Setting}
We randomly split our dataset into 80\% training, 10\% validation, and 10\% test sets. For subword information embeddings, we chose the number of dimensions of phonetic-level embedding as 20, and the number of dimensions of character-level embedding as 20. For word-level embeddings, the pre-trained FastText and BERT\_Base had 300 dimensions and 768 dimensions, respectively. To ensure consistency of all deep learning models (including baselines), we manually fixed the batch size as 128. Other than that, we applied grid search for determining the best hyperparameters for all models (including baselines). All weights/parameters of all models were fine-tuned to achieve each one's best result. We used ReLU as activation functions, cross-entropy as loss measurement, and Adam as the optimizer.

In the character-level attack scenario, all manipulated misspelling was created in only the test set, while we kept the training and validation sets without any change, avoiding any model seeing or remembering the adversarial attack. The proportion of manipulated data varies from 0\% to 100\%, at a step size of 10\%. We report the result of them in Section~\ref{subsec:bugAttack}.

\begin{table}[t]
    \centering
    \small
    \caption{Performance of our SWE2 models and baselines without the adversarial attack.}
    \vspace{-10pt}
    \scalebox{0.85}{
    \begin{tabular}{l | c c | c c} \hline
        \multirow{2}{*}{MODEL}&  Overall&  Macro& \multicolumn{1}{c}{Leg.}&    \multicolumn{1}{c}{Hate S.}\\
        & Acc.&    F1&    F1& F1\\ \hline
        Davidson'17&   .904& .764&  .946& .583\\
        Text-CNN'14&   .935& .894&  .960& .829\\
        Waseem'16&   .950& .913&  .970& .857\\
        Zhang'18&   .957& .927& .974&   .879\\
        Badjatiya'17&   .933& .892& .959& .826\\
        Fermi'19 SVM&   .821&   .740&   .885&   .595\\
        DirectBERT'19&   .942&   .902&  .965&    .839\\ \hline
        SWE2 w/ BERT&   \textbf{.975}&   \textbf{.953}& \textbf{.985}&   \textbf{.921}\\
        SWE2 w/ FastText5&   .974&   .950& .984&   .915\\ \hline
    \end{tabular}}
    \label{tab:noBug}
    \vspace{-10pt}
\end{table}

\vspace{-10pt}
\subsection{Experimental Results}
\subsubsection{\textbf{Performance under no adversarial attack}}
Table~\ref{tab:noBug} shows performance of our models (\emph{SWE2~w/~BERT} and \emph{SWE2~w/~FastText}) and baselines under no-attack scenario.
Our models outperformed all baseline models. In particular, \emph{SWE2~w/~BERT} achieved 0.975 accuracy and 0.953 macro F1. The macro F1 balances the performance of both classes as the dataset is not balanced.

In addition, one would observe that baseline models' performance in Hate Speech class is not as good as the Legitimate Class (at least 10\% gap between F1 in two classes). This is also reported in \cite{waseem2016hateful}, as they found detecting offensive language and legitimate speech is easier than hate speech detection in their experiments. Our model, however, did exceptionally well in the Hate Speech class in terms of F1 score. Especially, our \emph{SWE2~w/~BERT} achieved 0.921 F1, improving 4.8\% compared with the best baseline (i.e, \emph{Zhang'18}). Although \emph{SWE2~w/~BERT} is slightly better than \emph{SWE2~w/~FastText}, the performance of different embeddings does not differ much, showing contribution of our overall framework design.

\vspace{-5pt}
\subsubsection{\textbf{Performance under the adversarial attack}}
\label{subsec:bugAttack}

\begin{figure}[t]
    \centering
    \includegraphics[width=.65\linewidth]{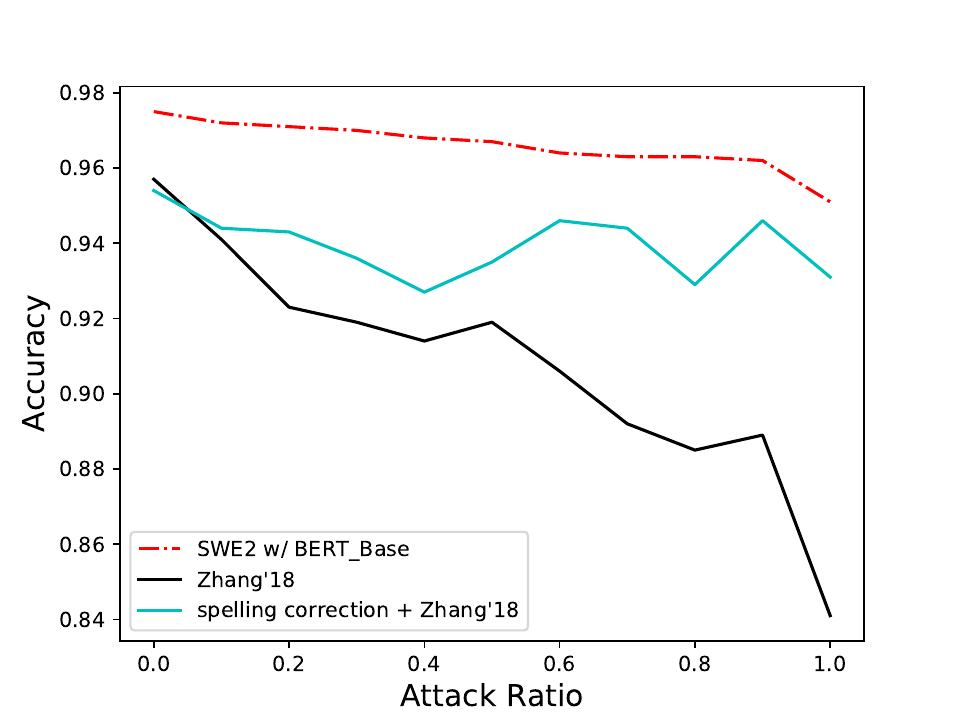}
    \vspace{-10pt}
    \caption{Accuracy of our SWE2 model and the best baseline under the adversarial attack.}
    \label{fig:Acc4model}
    \vspace{-12pt}
\end{figure}

To measure the robustness of our model under the character-level adversarial attack, we change an attack ratio from 0 to 1 (i.e., manipulating 0\% to 100\% messages in the test set). For example, 0.1 attack ratio means 10\% of hate speeches and 10\% of legitimate messages are manipulated by the attack model as described in Section~\ref{subsec:Manipulate}. We intentionally manipulated both hate speeches and legitimate messages to avoid possible bias that models learned to judge hate speech based on a higher ratio of unknown words rather than understanding actual hate speech content (manipulation actually changed the spellings of words, so they may lead to out of vocabulary (OOV) cases for some embeddings). Another important reason is that as we reported earlier, even legitimate users often make typos. We are interested to push the case to an extreme, to see whether models are still capable to tell the difference between legitimate speeches against hate speeches. As \emph{Zhang'18} was the best baseline in the previous experiment, we compare the performance of our model (\emph{SWE2~w/~BERT}) against it. In addition, we also applied auxiliary spelling correction tools~\cite{baziotis-pelekis-doulkeridis:2017:SemEval2} as an preprocessing technique into \emph{Zhang'18} to check the effectiveness of spelling correction under the adversarial attack.

Figure~\ref{fig:Acc4model} shows accuracy of our model and the best baseline under the adversarial attack (macro F1 has almost same pattern). We observe that our model consistently performed well under the adversarial attack. In particular, in the 50\% (0.5) attack ratio, our model still achieved high performance, returning 0.967 accuracy and 0.934 macro F1 score while \emph{Zhang'18} produced 0.919 accuracy and 0.887 macro F1. Even under 100\% attack ratio (very extreme scenario), our model achieved 0.951 accuracy and 0.902 macro F1, while \emph{Zhang'18} reached 0.841 accuracy and 0.783 macro F1. The performance of the baseline dropped rapidly as we increased an attack ratio.
Another observation is that spelling correction in general helped mitigating character level attacks to \emph{Zhang'18} but its performance was not stable. The spelling correction helped partly with some misspellings while not working in others. For example, given `this how you brign the city out url' text, it was then wrongly recovered as `this how you brain the city out url' (i.e., recovered `brign' into `brain' instead of `bring'). Overall, our model still outperformed \emph{Zhang'18} with and without spelling correction. 
We also tried manipulating only hate speeches in the test set without manipulating legitimate speeches. In the experiment, we saw the similar pattern in which our model consistently and robustly performed well, but the best baselines rapidly dropped its performance as increasing an attack ratio. These experiments confirm robustness of our model under the attack scenario.

\vspace{-5pt}
\subsubsection{\textbf{Ablation Study}}

\begin{table}[t]
    \centering
    \small
    \caption{Performance of ablation study.}
    \vspace{-10pt}
    \scalebox{0.85}{
    \begin{tabular}{l | c c | c c} \hline
        \multirow{2}{*}{MODEL}& \multicolumn{2}{c|}{Attack 0\%}& \multicolumn{2}{c}{Attack 50\%}\\
        &  Acc.& Macro F1& Acc.&    Macro F1\\ \hline
        SWE2 w/ BERT&   .975&   .953&   .966&   .934\\
        ~~--Char&  .959& .928&  .956&   .923\\
        ~~--Pho&  .960&   .931& 958&    .926\\
        ~~--Char\&Pho& .957&   .923&    .956&   .923\\
        ~~--LSTMs&    .940&   .863&    .915&   .821\\
        \hline
    \end{tabular}}
    \label{tab:Ablation}
    \vspace{-10pt}
\end{table}

We show the results of the ablation study in Table~\ref{tab:Ablation}. We used the BERT\_Base as the default word embedding method and tested the model's performance without certain parts to see their contribution. We show results of removing \emph{target} word's character embedding, phonetic embedding, both of them (i.e., no explicit \emph{target} word information is given), or our two two-layer LSTMs (i.e., no explicit global information and no implicit prediction for the \emph{target} word. It also means we remove/not use all words except the \emph{target} word in a message). In running these experiments, we not only show each part of embeddings' contribution but also examine the actual effect of the located \emph{target} word. Note that we only show the results of 0\% attack (i.e., no attack) and 50\% attack due to the limited space, but results of all other ratios are similar.

Under no attack scenario, character-level \emph{target} word embedding contributed 1.6\% accuracy and 2.5\% macro F1 improvement. Phonetic-level \emph{target} word embedding contributed 1.5\% accuracy and 2.2\% macro F1 improvement. Our LSTM architecture contributed 3.5\% accuracy and 9\% macro F1 improvement. Under the 50\% attack scenario, character-level \emph{target} word embedding contributed 1.0\% accuracy and 1.1\% macro F1 improvement. Phonetic-level \emph{target} word embedding contributed 0.8\% accuracy and 0.8\% macro F1 improvement. our LSTM architecture contributed 5.1\% accuracy and 11.3\% macro F1 improvement.

The results make sense. LSTMs function as the backbone of the framework, as losing LSTMs deprives most semantics as well as all synthetic information from the sentence. In other words, keeping only one word from a sentence and eliminating the other words are not sufficient to judge whether a speech is legitimate or not. On the other hand, dropping the target word would deprive the advantage of our framework, so it makes our model's accuracy lower. Overall, all of the components in our framework positively contributed in terms of improving the performance and keeping the robustness under the adversarial attack.

\begin{table}[t]
    \centering
    \small
    \caption{SWE2~w/~BERT under various class ratios.}
    \vspace{-10pt}
    \scalebox{0.85}{
    \begin{tabular}{c | c c | c c} \hline
        \multirow{2}{*}{Leg.:Hate S.}&    Overall&    Macro& Leg.& Hate S.\\
        & Acc.&    F1&  F1& F1\\ \hline
        1:1&   .953&   .953&    .953&   .953\\
        2:1&   .961&   .956&    .970&   .941\\
        3:1&   .965&   .953&    .977&   .930\\
        4:1&   .972&   .958&    .983&   .930\\
        5:1&   .975&   .953&    .985&   .921\\
        \hline
    \end{tabular}}
    \label{tab:ClassRatio}
    \vspace{-10pt}
\end{table}

\vspace{-5pt}
\subsection{Varying a class ratio}
To understand how a class ratio in the dataset affects performance of our model (as an additional experiment to measure robustness of our model), we varied a class ratio of the dataset, by downsampling legitimate messages. In particular, we tested a ratio from 1:1 to 5:1. 5:1 ratio means the original dataset without downsampling.

Table~\ref{tab:ClassRatio} shows performance of our model under various class ratios. As we increase the number of legitimate messages (from 1:1 to 5:1), accuracy has increased but macro F1 has been consistent. The result makes sense since our models were trained with different respective class weights, depending on a class ratio (i.e., if more legitimate messages are in a dataset, the model will try not to misclassify the legitimate messages in the training process). The consistent macro F1 indicates the robustness of our model/framework regardless of a class ratio. In practice, since there would be more legitimate messages, our model would achieve better accuracy and similar macro F1.

\vspace{-5pt}
\section{Discussion and Analysis}
\label{sec:analysis}
\subsection{How many hate speech words do hate speeches contain?}
We further investigate the hate speeches distribution to answer the following questions:
\squishlist
\item Do all hate speeches contain some hate speech words?
\item As our framework puts extra emphasis on one most significant word (i.e., \emph{target} word) in each speech/message, can it handle speeches with no hate speech word or speeches with multiple hate speech words?
\squishend
We used the hate speech keyword dictionary provided by \cite{elsherief2018hate} to identify hate speech words in hate speeches of our dataset. Out of 16,703 hate speeches in our dataset, 12,378 hate speeches contained hate speech words. Among them, 11,835 hate speeches contained only one hate speech word; 443 hate speeches contained multiple hate speech words. The remaining 4,425 hate speeches did not contain any hate speech word.

Our \emph{SWE2~w/~BERT} detected 69.0\%, 99.1\%, 100\% correctly for hate speeches without hate speech word, hate speeches with single hate speech word, and hate speeches with multiple hate speech words, respectively. Although it is generally hard to detect hate speeches without any specific hate speech word, our model still reasonably identified them. On the other side, our model did exceptionally well in detecting hate speeches with one or multiple hate speech words, indicating that our model is effective in detecting hate speeches.

\vspace{-10pt}
\subsection{Why choose the most significant word?}
\label{sec:targetword}
We reason the choice of focusing the one most significant word with the following facts from different perspectives:

\squishlist
    \item \textbf{Only focusing on random words does not work.} Previous work in \cite{li2019textbugger} showed that random attack on words is not successful. A successful attack has to choose certain words rather than randomly chosen words to attack. Consequently, focusing on random words to defend attacks is also not useful.
    \item \textbf{The most important word contributes most in both sentiment and semantic meaning of the message.} Figure~\ref{fig:sentiment} shows how the most significant word almost dominated a sentiment score of each message. In addition, we observe that some hate speech posters tend to manipulate the important word to evade existing detection approaches.
    \item \textbf{Focusing on one single most important word succeeded with detecting hate speech with multiple hate speech words or without hate speech word.} As we analyzed before, there are many hate speeches with multiple hate speech words in the dataset, and also over four thousand hate speeches without any specific hate speech word. However, the strategy of focusing on the single most important word still succeeded to detect hate speeches correctly overall.
    \item \textbf{We show a real evidence in Case Study, indicating the necessity of the focusing strategy.} Section \ref{subsec:CaseStudy} shows a real hate speech that was correctly predicted by our model, but the best baseline misclassified it to a legitimate message because only the most important word in the message was strongly negative while the other words were positive. The best baseline failed to put extra emphasis on the most important word. Thus other words' meanings mitigated the impact of the \emph{target} word, eventually leading to misclassification.
\squishend

\vspace{-5pt}
\subsection{How is sentiment of a message changed under character-level manipulation?}
\label{subsec:sentiment}

We try to reason and give out evidence of why sentiment-based features are vulnerable against character-level manipulation/adversarial attack. As sentiment-based features' atomized elements are words, the manipulated words lost their sentiment significance. Thus, it will affect the actual sentiment measure of each message.

\begin{figure}[t]
    \centering
    \includegraphics[width=\linewidth,height=1.5in]{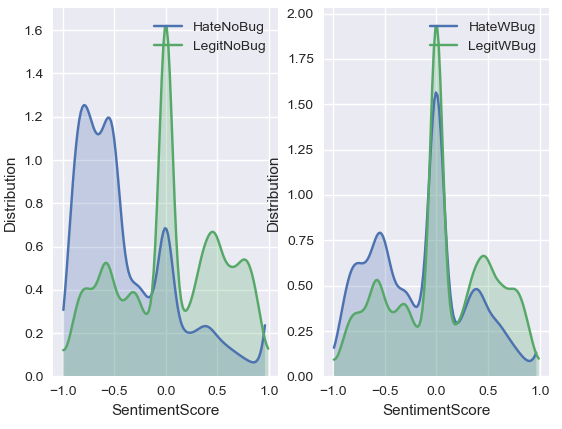}
    \vspace{-20pt}
    \caption{Sentiment score before and after the character-level manipulation.}
    \label{fig:sentiment}
    \vspace{-17pt}
\end{figure}

To prove the correctness of the hypothesis, we used VADER \cite{hutto2014vader} as the sentiment analysis tool and showed a sentiment change between before and after the character-level manipulation in Figure~\ref{fig:sentiment}. For each given message, we generate the compound sentiment score generated by the VADER. Such a score varies in a range of $[-1,1]$. The closer the score gets to 1 indicates the message is more sentimentally positive, while the closer the score gets to \emph{-1} means the message is more sentimentally negative. In the left subfigure, we show sentiment score distribution of hate speech and legitimate messages before applying the character-level manipulation in our dataset. The curve reflects the kernel density estimation of the given population distribution observations. In the right subfigure, we show the sentiment score distribution of hate speech and legitimate messages after applying the character-level manipulation.

We observe that sentiment based features contribute to separating hate speech against legitimate speech under no attack/manipulation in the left subfigure. However, they lost power when manipulations occur in the right subfigure. This result explains why the baselines using sentiment features (e.g., \emph{Davidson'17}) experienced a performance drop when they face the character-level adversarial attack. This figure also points out two interesting facts: 1) The sentiment does not change much for legitimate speech with the manipulated misspellings; and 2) The most significant word contributes the most in each speech's overall sentiment score.

\vspace{-5pt}
\subsection{\textbf{Humans vs. machine learning models under the character-level adversarial attack}}
\label{subsec:human}
Why can humans still correctly perceive the meaning of a message with typos, but machine learning models consider it as a hard problem\footnote{\url{https://en.wikipedia.org/wiki/Typoglycemia}}? The way humans memorize words is different from machine learning models in three ways:
\squishlist
    \item Humans' reading is sequential, and they can tolerate some errors/typos\footnote{\url{https://bit.ly/33zupxw} and \url{https://bit.ly/3idUXIR}}, while usually machine learning models compare strings of characters, so the result is strictly boolean, which excludes any tolerance.

    \item When reading, humans encode the information they receive mainly from eyes, graphically recognizing words. Thus they can recognize similar characters\footnote{\url{https://bit.ly/33wCHX9}} such as `s' and `\$', or `l' and `1'. While inside the machine learning models, every character is encoded differently and independently, thus there is no such correlation between different characters.

    \item Humans incorporate more side information than machine learning models during the reading and memorizing. They not only read words quietly but also explicitly or unintentionally associate pronunciations with the words \cite{toutanova2001pronunciation}. Thus, the words `jews' and `jooz' can be recognized as if they are the same thing even when we never actually learned `jooz' before. However, if machine learning models do not have the term `jooz' in their vocabulary, they would treat the word as an unknown word, failing to extract useful information from it.
\squishend

Even though some hate speech messages are manipulated, the ultimate goal of these hate speech messages will never be changed. In other words, they have to be understood by humans, to make an actual impact on society. Based on the reasoning and analysis, in our models, we incorporated both word-level and subword level information not only to enrich the content of a message but also help them better recognize/recover the original meaning of the message. Our experimental results in the previous section confirmed the effectiveness of our framework and models, achieving high hate speech detection accuracy and F1.

\vspace{-5pt}
\subsection{Case Study}
\label{subsec:CaseStudy}
In this subsection, we conduct a micro-scale case study, where we show examples\footnote{All examples in the case study belong to the public datasets released by other researchers described in Section \ref{sec:Experiment}.} that our models correctly predicted and that our models misclassified. By doing this analysis, we can shed some light on designing better hate speech detection approaches. For convenience, we compare our \emph{SWE2~w/~BERT} against the best performing baseline -- \emph{Zhang'18}.

\vspace{-5pt}
\subsubsection{\textbf{Advantage Analysis}}
\label{subsec:advantage}
We first show two examples randomly sampled from instances that were correctly predicted by our model, while the best baseline misclassified it.
\squishlist
\item An example without the adversarial attack: \textit{\textbf{``... win the faggot award congrats ...''}}
\squishend

When we read this message, we can find that all the words except `faggot' are positive words. Thus, the baseline does not emphasize the most important word, `faggot', which does not have enough influence on the overall sentiment score. Thus, its negative impact is mitigated by its surrounding positive words, leading to misclassification. However, in our model, `faggot', is recognized as an important keyword. It puts extra attention and emphasis on the important word. The character-level information is captured by our model and further guarantees correct prediction.
\squishlist
\item A real-world example before applying the adversarial attack: \textit{\textbf{``... look like a redneck ... confederate flag tattoo on ... ass.''}}
\item The same example after applying the adversarial attack: \textit{\textbf{``... look like a rednecj ... confederate flag tattoo on ... ass.''}}
\squishend

By comparing the above examples, we noticed that `rednecj' was manipulated from `redneck' by the \emph{sub-c} manipulation. In the baseline, this misspelled word is roughly identified as an unknown word. However, our model still learns features from it by using character embedding and phonetic embedding, mimicking how humans read it as described in Section~\ref{subsec:human}.

\vspace{-5pt}
\subsubsection{\textbf{Error Analysis}}
Next, we show an example message that even our model did not correctly identify.

\squishlist
\item \textit{\textbf{``... fucking hate you ... but thank you ... dick van dyke.''}}
\squishend

The example was manually labeled as a legitimate tweet \cite{waseem2016you} but was misclassified by both of our model and the best baseline. The message itself contains several aggressive and sentimentally negative words such as `fucking' and `dick'. Besides, it has `dyke', which is listed as a hate speech keyword. However, the real meaning is to express thanks in a joking way. It is still challenging for our model to further deeply understand and detect the text with complicated sentiment and mood swings. Using better domain-specific embedding may help a model to predict its label correctly.

\vspace{-5pt}
\section{Conclusion}
\label{sec:Conclusion}
In this paper, we proposed a novel hate speech detection framework that incorporates subword information and word-level semantic information. By addressing the importance of successfully identifying manipulated words and focusing on the most significant word in the message, and by using attention mechanisms to extract side information, our models outperformed all 7 state-of-the-art baselines. Under no attack, our \emph{SWE2~/w~BERT} achieved 0.975 accuracy and 0.953 macro F1, and under 50\% attack, our \emph{SWE2~/w~BERT} still achieved 0.967 accuracy and 0.934 macro F1, showing effectiveness and robustness of our model.

\vspace{-5pt}
\begin{acks}
This work was supported in part by NSF grant CNS-1755536, AWS Cloud Credits for Research, and Google Cloud. 
\end{acks}

\bibliographystyle{ACM-Reference-Format}
\bibliography{sample-base}

\end{document}